\documentclass{article}
\usepackage{colacl98}
\usepackage{epsfig}
\usepackage{rotating}
\usepackage{nobi}
\usepackage{sap}
\usepackage{nobiDG}

\newlength{\oldtabcolsep}
\begin{document}


\title{	\appearedIn{In: Proc. of ANLP--NAACL, Apr 29 -- May 4, 2000, pp.ANLP-325-330}{325}
	Improving Testsuites via Instrumentation}

\author{Norbert Br\"{o}ker \\
	Eschenweg 3 \\
	69231 Rauenberg \\
	Germany\\
	\texttt{norbert.broeker@sap.com}}


\maketitle

\begin{abstract}
This paper explores the usefulness of a technique from software
engineering, namely code instrumentation, for the development of
large-scale natural language grammars. Information about the usage of
grammar rules in test sentences is used to detect untested rules, redundant
test sentences, and likely causes of overgeneration. Results show that less
than half of a large-coverage grammar for German is actually tested by
two large testsuites, and that 10--30\% of testing time is redundant. The
methodology applied can be seen as a re-use of grammar writing knowledge
for testsuite compilation.
\end{abstract}

\section{Introduction}
	\lbl{ch:intro}

Computational Linguistics (CL) has moved towards the marketplace: One finds
programs employing CL-techniques in every software shop: Speech
Recognition, Grammar and Style Checking, and even Machine Translation are
available as products. While this demonstrates the applicability of the
research done, it also calls for a rigorous development methodology of such
CL application products.


In this paper,%
\footnote{%
The work reported here was conducted during my time at the Institut f\"{u}r
Maschinelle Sprachverarbeitung (IMS), Stuttgart University, Germany.
}%
I describe the adaptation of a technique from Software
Engineering, namely code instrumentation, to grammar development.
Instrumentation is based on the simple idea of marking any piece of code
used in processing, and evaluating this usage information afterwards. The
application I present here is the evaluation and improvement of grammar and
testsuites; other applications are possible.

\subsection{Software Engineering vs. Grammar Engineering}

Both software and grammar development are similar processes: They result in
a system transforming some input into some output, based on a functional
specification (e.g., cf. \cite{Ciravegna1998} for the application of a
particular software design methodology to linguistic engineering). Although
Grammar Engineering usually is not based on concrete specifications,
research from linguistics provides an informal specification.

Software Engineering developed many methods to assess the quality of a
program, ranging from static analysis of the program code to dynamic
testing of the program's behavior. Here, we adapt dynamic testing, which
means running the implemented program against a set of test cases. The test
cases are designed to maximize the probability of detecting errors in the
program, i.e., incorrect conditions, incompatible assumptions on subsequent
branches, etc. (for overviews, cf. \cite{Hetzel1988,Liggesmeyer1990}).




\subsection{Instrumentation in Grammar Engineering}

How can we fruitfully apply the idea of measuring the coverage of a set of
test cases to grammar development? I argue that by exploring the relation
between grammar and testsuite, one can improve both of them. Even the
traditional usage of testsuites to indicate grammar gaps or overgeneration
can profit from a precise indication of the grammar rules used to parse the
sentences (cf. Sec.\ref{ch:negative}). Conversely, one may use the grammar
to improve the testsuite, both in terms of its coverage (cf.
Sec.\ref{ch:suite-complete}) and its economy (cf. Sec.\ref{ch:economic}).

Viewed this way, testsuite writing can benefit from grammar development
because both describe the syntactic constructions of a natural language.
Testsuites systematically list these constructions, while grammars give
generative procedures to construct them. Since there are currently many
more grammars than testsuites, we may re-use the work that has gone into
the grammars for the improvement of testsuites.

The work reported here is situated in a large cooperative project aiming at
the development of large-coverage grammars for three languages. The
grammars have been developed over years by different people, which makes
the existence of tools for navigation, testing, and documentation
mandatory. Although the sample rules given below are in the format of LFG,
nothing of the methodology relies on the choice of linguistic or
computational paradigm.

\section{Grammar Instrumentation}
	\lbl{ch:grammar}

Measures from Software Engineering cannot be simply transferred to Grammar
Engineering, because the structure of programs is different from that of
unification grammars. Nevertheless, the \emph{structure} of a grammar
allows the derivation of suitable measures, similar to the structure of
programs; this is discussed in Sec.\ref{ch:gramm-crit}. The actual
instrumentation of the grammar depends on the formalism used, and is
discussed in Sec.\ref{ch:gramm-instr}.

\subsection{Coverage Criteria}
	\lbl{ch:gramm-crit}

\begin{figure}[t]
\centering
\begin{tabular}{llcl}
VP\prod	& V	& \mcl{2}{$\downarrow = \uparrow$;} \\
	& NP?	& \mcl{2}{$\downarrow = (\uparrow \mbox{OBJ})$;} \\
	& PP*	& \{	& $\downarrow = (\uparrow \mbox{OBL})$; \\
	&	& \|	& $\downarrow \in (\uparrow \mbox{ADJUNCT})$;\}.
\end{tabular}
\caption{Sample Rule 
	\label{fig:rule}}
\end{figure}

Consider the LFG grammar rule in Fig.~\ref{fig:rule}.%
\footnote{
Notation: \texttt{?}/\texttt{*}/\texttt{+} represent optionality/iteration
including/excluding zero occurrences on categories. Annotations 
to a category specify equality (\texttt{=}) or set membership ($\in$) of
feature values, or non-existence of features ($\neg$); they are terminated
by a semicolon (\texttt{;}). Disjunctions are given in braces
(\texttt{\{...\|...\}}). $\uparrow$ ($\downarrow$) are metavariables
representing the feature structure corresponding to the mother (daughter)
of the rule. Comments are enclosed in quotation marks (\texttt{"..."}). Cf.
\cite{Kaplan+Bresnan1982} for an introduction to LFG notation. 
}
On first view, one could require of a testsuite that each such rule is
exercised at least once. Further thought will indicate that there are
hidden alternatives, namely the optionality of the NP and the PP. The rule
can only be said to be thoroughly tested if test cases exist which test
both presence and absence of optional constituents (requiring 4 test cases
for this rule).

In addition to context-free rules, unification grammars contain equations
of various sorts, as illustrated in Fig.\ref{fig:rule}. Since these
annotations may also contain disjunctions, a testsuite with complete rule
coverage is not guaranteed to exercise all equation alternatives. The
phrase-structure-based criterion defined above must be refined to cover all
equation alternatives in the rule (requiring two test cases for the PP
annotation). Even if we assume that (as, e.g., in LFG) there is at least
one equation associated with each constituent, equation coverage does not
subsume rule coverage: Optional constituents introduce a rule disjunct
(without the constituent) that is not characterizable by an equation. A
measure might thus be defined as follows:

\begin{description}
\item[disjunct coverage] The disjunct coverage of a testsuite is the
quotient
\[
T_{\mbox{dis}}=\frac{\mbox{number of disjuncts tested}}{\mbox{number of disjuncts in grammar}} 
\]
where a disjunct is either a phrase-structure alternative, or an annotation
alternative. Optional constituents (and equations, if the formalism allows
them) have to be treated as a disjunction of the constituent and an empty
category (cf. the instrumented rule in Fig.\ref{fig:instr-example} for an
example). 
\end{description}

\noindent
Instead of considering disjuncts in isolation, one might take their
interaction into account. The most complete test criterion, doing this to
the fullest extent possible, can be defined as follows:

\begin{description}
\item[interaction coverage] The interaction coverage of a testsuite is the
quotient
\[
T_{\mbox{inter}}=\frac{\mbox{number of disjunct combinations tested}}{\mbox{number of legal disjunct combinations}} 
\]

There are methodological problems in this criterion, however. First, the
set of legal combinations may not be easily definable, due to far-reaching
dependencies between disjuncts in different rules, and second, recursion
leads to infinitely many legal disjunct combinations as soon as we take the
number of usages of a disjunct into account. Requiring complete interaction
coverage is infeasible in practice, similar to the path coverage criterion
in Software Engineering. 
\end{description}


We will say that an analysis (and the sentence receiving this analysis)
\emph{relies on} a grammar disjunct if this disjunct was used in
constructing the analysis.

\subsection{Instrumentation}
	\lbl{ch:gramm-instr}

Basically, grammar instrumentation is identical to program instrumentation:
For each disjunct in a given source grammar, we add grammar code that will
identify this disjunct in the solution produced, iff that disjunct has been
used in constructing the solution.

Assuming a unique numbering of disjuncts, an annotation of the form {\tt
DISJUNCT-nn~=~+} can be used for marking. To determine whether a certain
disjunct was used in constructing a solution, one only needs to check
whether the associated feature occurs (at some level of embedding) in the
solution. Alternatively, if set-valued features are available, one can use
a set-valued feature {\tt DISJUNCTS} to collect atomic symbols representing
one disjunct each: {\tt DISJUNCT-nn~$\in$~DISJUNCTS}.

One restriction is imposed by using the unification formalism, though: One
occurrence of the mark cannot be distinguished from two occurrences, since
the second application of the equation introduces no new information. The
markers merely unify, and there is no way of counting.

Therefore, we have used a special feature of our grammar development
environment: Following the LFG spirit of different representation levels
associated with each solution (so-called \textit{projections}), it provides
for a multiset of symbols associated with the complete solution, where
structural embedding plays no role (so-called \textit{optimality
projection}; see \cite{Frank1998}). In this way, from the root node of each
solution the set of all disjuncts used can be collected, together with a
usage count.

Fig.~\ref{fig:instr-example} shows the rule from Fig.\ref{fig:rule} with
such an instrumentation; equations of the form \texttt{DISJUNCT-nn}$ \in
o*$ express membership of the disjunct-specific atom \texttt{DISJUNCT-nn}
in the sentence's multiset of disjunct markers. 

\begin{figure}[t]
\centering
\small
\begin{tabular}{llcl}
VP\prod	& V	& \mcl{2}{$\downarrow = \uparrow$;} \\
	& \{ e	& \mcl{2}{DISJUNCT-001 $\in o*$;} \\
	& \| NP	& \mcl{2}{$\downarrow = (\uparrow \mbox{OBJ})$} \\
	&	& \mcl{2}{DISJUNCT-002 $\in o*$;} \\
	& \}	& \\
	& \{ e	& \mcl{2}{DISJUNCT-003 $\in o*$;} \\
	& \| PP+& \{	& $\downarrow = (\uparrow \mbox{OBL})$ \\
	&	&	& DISJUNCT-004 $\in o*$; \\
	&	& \|	& $\downarrow \in (\uparrow \mbox{ADJUNCT})$ \\
	&	&	& DISJUNCT-005 $\in o*$;\}. \\
	& \}	& \\
\end{tabular}
\caption{Instrumented rule
	\label{fig:instr-example}}
\end{figure}

\subsection{Processing Tools}

Tool support is mandatory for a scenario such as instrumentation: Nobody
will manually add equations such as those in Fig.~\ref{fig:instr-example}
to several hundred rules. Based on the format of the grammar rules, an
algorithm instrumenting a grammar can be written down easily. 

Given a grammar and a testsuite or corpus to compare, first an instrumented
grammar must be constructed using such an algorithm. This instrumented
grammar is then used to parse the testsuite, yielding a set of solutions
associated with information about usage of grammar disjuncts. Up to this
point, the process is completely automatic. The following two sections
discuss two possibilities to evaluate this information.

\section{Quality of Testsuites}
	\lbl{ch:suites}

This section addresses the aspects of completeness (``does the testsuite
exercise all disjuncts in the grammar?'') and economy of a testsuite (``is
it minimal?'').

Complementing other work on testsuite construction (cf.
Sec.\ref{ch:compare-suites}), we will assume that a grammar is already
available, and that a testsuite has to be constructed or extended. While
one may argue that grammar and testsuite should be developed in parallel,
such that the coding of a new grammar disjunct is accompanied by the
addition of suitable test cases, and vice versa, this is seldom the case.
Apart from the existence of grammars which lack a testsuite, and to which
this procedure could be usefully applied, there is the more principled
obstacle of the evolution of the grammar, leading to states where
previously necessary rules silently loose their usefulness, because their
function is taken over by some other rules, structured differently. This is
detectable by instrumentation, as discussed in Sec.\ref{ch:suite-complete}.

On the other hand, once there is a testsuite, you want to use it in the
most economic way, avoiding redundant tests. Sec.\ref{ch:economic} shows
that there are different levels of redundancy in a testsuite, dependent on
the specific grammar used. Reduction of this redundancy can speed up the
test activity, and give a clearer picture of the grammar's performance.

\subsection{Testsuite Completeness}
	\lbl{ch:suite-complete}

If the disjunct coverage of a testsuite is 1 for some grammar, the
testsuite is \emph{complete} w.r.t. this grammar. Such a testsuite can
reliably be used to monitor changes in the grammar: Any reduction in the
grammar's coverage will show up in the failure of some test case (for
negative test cases, cf. Sec.\ref{ch:negative}).

If there is no complete testsuite, one can -- via instrumentation --
identify disjuncts in the grammar for which no test case exists. There
might be either (i) appropriate, but untested, disjuncts calling for the
addition of a test case, or (ii) inappropriate disjuncts, for which one
cannot construct a grammatical test case relying on them (e.g., left-overs
from rearranging the grammar). Grammar instrumentation singles out all
untested disjuncts automatically, but cases (i) and (ii) have to be
distinguished manually.

Checking completeness of our local testsuite of 1787 items, we found that
only 1456 out of 3730 grammar disjuncts in our German grammar were tested,
yielding $T_{dis} = 0.39$ (the TSNLP testsuite containing 1093 items tests
only 1081 disjuncts, yielding $T_{dis} = 0.28$).%
\footnote{%
There are, of course, unparsed but grammatical test cases in both
testsuites, which have not been taken into account in these figures. This
explains the difference to the overall number of 1582 items in the German
TSNLP testsuite. 
} 
Fig.\ref{fig:gap} shows an example of a gap in our testsuite (there are no
examples of circumpositions), while Fig.\ref{fig:inappropriate} shows an
inapproppriate disjunct thus discovered (the category ADVadj has been
eliminated in the lexicon, but not in all rules). Another error class is
illustrated by Fig.\ref{fig:inconsistent}, which shows a rule that can
never be used due to an LFG coherence violation; the grammar is
inconsistent here.%
\footnote{%
Test cases using a free dative pronoun may be in the testsuite, but receive
no analysis since the grammatical function FREEDAT is not defined as such
in the configuration section.
}

\begin{figure}[t]
\centering
\small
\begin{tabular}{lll}
PPstd\prod	& Pprae		& $\downarrow = \uparrow$;\\
		& NPstd		& $\downarrow = (\uparrow \mbox{OBJ})$; \\
		& \{ e		& DISJUNCT-011 $\in o*$; \\
		& \| Pcircum	& $\downarrow = \uparrow$;\\
		&		& DISJUNCT-012 $\in o*$ \\
		&		& "unused disjunct"; \\
		& \}		& \\
\end{tabular}
\caption{Appropriate untested disjunct}
	\label{fig:gap}
\end{figure}

\begin{figure}[t]
\centering
\small
\begin{tabular}{lcll}
ADVP\prod	& \{	& \{ e		& DISJUNCT-021 $\in o*$;\\
		&	& \| ADVadj	& $\downarrow = \uparrow$ \\
		&	&		& DISJUNCT-022 $\in o*$ \\
		&	&		& "unused disjunct"; \\
		&	& \} \\
		& 	& ADVstd	& $\downarrow = \uparrow$ \\
		&	&		& DISJUNCT-023 $\in o*$ \\
		&	&		& "unused disjunct"; \\
		&	& \} \\
		& \| 	& ... \\
		& \}. \\
\end{tabular}
%
\caption{Inappropriate disjunct
	\label{fig:inappropriate}}
\end{figure}

\begin{figure}[t]
\centering
\small
\begin{tabular}{lcll}
VPargs \prod	& \{	& ... \\
		& \|	& PRONstd	& $\downarrow = (\uparrow \mbox{FREEDAT})$ \\
		&	&		& $(\downarrow \mbox{CASE}) = \mbox{dat}$ \\
		&	&		& $(\downarrow \mbox{PRON-TYPE}) = \mbox{pers}$ \\
		&	&		& $\neg(\uparrow \mbox{OBJ2})$ \\
		&	&		& DISJUNCT-041 $\in o*$ \\
		&	&		& "unused disjunct"; \\
		& \|	& ... \\
		& \}. \\
\end{tabular}
\caption{Inconsistent disjunct}
	\label{fig:inconsistent}
\end{figure}

\subsection{Testsuite Economy}
	\lbl{ch:economic}

Besides being complete, a testsuite must be economical, i.e., contain as
few items as possible without sacrificing its diagnostic capabilities.
Instrumentation can identify redundant test cases. Three criteria can be
applied in determining whether a test case is redundant:

\begin{description}
\item[similarity] There is a set of other test cases which jointly rely on
all disjunct on which the test case under consideration relies.

\item[equivalence] There is a single test case which relies on exactly the
same combination(s) of disjuncts.

\item[strict equivalence] There is a single test case which is equivalent
to and, additionally, relies on the disjuncts exactly as often as, the test
case under consideration.
\end{description}

For all criteria, lexical and structural ambiguities must be taken into
account. Fig.\ref{fig:equivs} shows some equivalent test cases derived from
our testsuite: Example 1 illustrates the distinction between equivalence
and strict equivalence; the test cases contain different numbers of
attributive adjectives, but are nevertheless considered equivalent. Example
2 shows that our grammar does not make any distinction between adverbial
usage and secondary (subject or object) predication. Example 3 shows test
cases which should not be considered equivalent, and is discussed below. 

\setlength{\tabcolsep}{\oldtabcolsep}
\begin{figure}[t]
\centering
\begin{tabular}{ll}
1	& ein guter alter Wein \\
	& ein guter alter trockener Wein \\
	& `\emph{a good old (dry) wine}' \\
2	& Er i\ss t das Schnitzel roh. \\
	& Er i\ss t das Schnitzel nackt. \\
	& Er i\ss t das Schnitzel schnell. \\
	& `\emph{He eats the schnitzel naked/raw/quickly.}' \\
3	& Otto versucht oft zu lachen. \\
	& Otto versucht zu lachen. \\
	& `\emph{Otto (often) tries to laugh.}' \\
\end{tabular}
\caption{Sets of equivalent test cases
	\label{fig:equivs}}
\end{figure}

The reduction we achieved in size and processing time is shown in
Table~\ref{tbl:reduced-suites}, which contains measurements for a test run
containing only the parseable test cases, one without equivalent test cases
(for every set of equivalent test cases, one was arbitrarily selected), and
one without similar test cases. The last was constructed using a simple
heuristic: Starting with the sentence relying on the most disjuncts,
working towards sentences relying on fewer disjuncts, a sentence was
selected only if it relied on a disjunct on which no previously selected
sentence relied. Assuming that a disjunct working correctly once will work
correctly more than once, we did not consider strict equivalence.

\begin{table}
\centering
\begin{tabular}{@{}l@{}rrrrr}
			& \begin{turn}{90}\shortstack{\# test cases}\end{turn} 
				& \begin{turn}{90}\shortstack{relative \\ size}\end{turn}
					& \begin{turn}{90}\shortstack{total \\ runtime (sec)}\end{turn} 
						& \begin{turn}{90}\shortstack{relative \\ runtime}\end{turn}
							& \begin{turn}{90}\shortstack{\# disjuncts \\ in grammar}\end{turn} \\
\hline
\multicolumn{6}{c}{TSNLP testsuite} \\
\hline
parseable		& 1093	& 100\%	& 1537	& 100\%	& 3561 \\ 	
no equivalents		& 783	& 71\%	& 665.3	& 43\%	& \\ 		
no similar cases	& 214	& 19\%	& 128.5	& 8\%	& \\		
\hline
\multicolumn{6}{c}{local testsuite} \\
\hline
parseable		& 1787	& 100\%	& 1213	& 100\%	& 5480 \\	
no equivalents		& 1600	& 89\%	& 899.5	& 74\%	& \\ 		
no similar cases	& 331	& 18\%	& 175.0	& 14\%	& \\ 		
\hline
\end{tabular}
\caption{Reduction of Testsuites
	\label{tbl:reduced-suites}}
\end{table}

We envisage the following use of this redundancy detection: There clearly
are linguistic reasons to distinguish all test cases in example 2, so they
cannot simply be deleted from the testsuite. Rather, their equivalence
indicates that the grammar is not yet perfect (or never will be, if it
remains purely syntactic). Such equivalences could be interpreted as a
reminder which linguistic distinctions need to be incorporated into the
grammar. Thus, this level of redundancy may drive your grammar development
agenda. The level of equivalence can be taken as a limited interaction
test: These test cases represent one complete selection of grammar
disjuncts, and (given the grammar) there is nothing we can gain by checking
a test case if an equivalent one was tested. Thus, this level of redundancy
may be used for ensuring the quality of grammar changes prior to their
incorporation into the production version of the grammar. The level of
similarity contains much less test cases, and does not test any
(systematic) interaction between disjuncts. Thus, it may be used during
development as a quick rule-of-thumb procedure detecting serious errors
only.

Coming back to example 3 in Fig.\ref{fig:equivs}, building equivalence
classes also helps in detecting grammar errors: If, according to the
grammar, two cases are equivalent which actually aren't, the grammar is
incorrect. Example 3 shows two test cases which are syntactically different
in that the first contains the adverbial \emph{oft}, while the other
doesn't. The reason why they are equivalent is an incorrect rule that
assigns an incorrect reading to the second test case, where the infinitival
particle \os{zu} functions as an adverbial.

\section{Negative Test Cases}
	\lbl{ch:negative}

To control overgeneration, appropriately marked ungrammatical sentences are
important in every testsuite. Instrumentation as proposed here only looks
at successful parses, but can still be applied in this context: If an
ungrammatical test case receives an analysis, instrumentation informs us
about the disjuncts used in the incorrect analysis. One (or more) of these
disjuncts must be incorrect, or the sentence would not have received a
solution. We exploit this information by accumulation across the entire
test suite, looking for disjuncts that appear in unusually high proportion
in parseable ungrammatical test cases.

In this manner, six grammar disjuncts are singled out by the parseable
ungrammatical test cases in the TSNLP testsuite. The most prominent
disjunct appears in 26 sentences (listed in Fig.\ref{fig:negative}), of
which group 1 is really grammatical and the rest fall into two
groups: A partial VP with object NP, interpreted as an imperative
sentence (group 2), and a weird interaction with the tokenizer
incorrectly handling capitalization (group 3).

\begin{figure}[t]
\begin{tabular}{@{}l@{ }ll@{ }l}
1	& Der Test f\"{a}llt leicht .	& 2	& Dieselbe schlafen . \\ 
	& Die schlafen .		& 	& Das schlafen . \\      
	&				&	& Eines schlafen . \\    
3	& Man schlafen .         	&	& Jede schlafen . \\     
	& Dieser schlafen .		& 	& Dieses schlafen . \\   
	& Ich schlafen .		& 	& Eine schlafen . \\     
	& Der schlafen .		& 	& Meins schlafen . \\    
	& Jeder schlafen .		& 	& Dasjenige schlafen . \\
	& Derjenige schlafen .		& 	& Jedes schlafen . \\    
	& Jener schlafen .		& 	& Diejenige schlafen . \\
	& Keiner schlafen .		& 	& Jenes schlafen . \\    
	& Derselbe schlafen .		& 	& Keines schlafen . \\   
	& Er schlafen .			& 	& Dasselbe schlafen . \\ 
	& Irgendjemand schlafen . \\
\end{tabular}
\caption{Sentences relying on a suspicious disjunct
	\label{fig:negative}}
\end{figure}

Far from being conclusive, the similarity of these sentences derived from a
suspicious grammar disjunct, and the clear relation of the sentences to
only two exactly specifiable grammar errors make it plausible that this
approach is very promising in reducing overgeneration.

\section{Other Approaches to Testsuite Construction}
	\lbl{ch:compare-suites}

Although there are a number of efforts to construct reusable large-coverage
testsuites, none has to my knowledge explored how existing grammars could
be used for this purpose.

Starting with \cite{Flickinger1987}, testsuites have been drawn up from a
linguistic viewpoint, \emph{``informed by [the] study of linguistics and
[reflecting] the grammatical issues that linguists have concerned
themselves with''} \cite[, p.4]{Flickinger1987}. Although the question is
not explicitly addressed in \cite{TSNLP-WP2.2}, all the testsuites reviewed
there also seem to follow the same methodology. The TSNLP project
\cite{Lehmann1996} and its successor DiET \cite{Netter1998}, which built
large multilingual testsuites, likewise fall into this category.

The use of corpora (with various levels of annotation) has been studied,
but even here the recommendations are that much manual work is required to
turn corpus examples into test cases (e.g., \cite{TSNLP-WP5.2}). The reason
given is that corpus sentences neither contain linguistic phenomena in
isolation, nor do they contain systematic variation. Corpora thus are used
only as an inspiration.

\cite{Oepen1998} stress the interdependence between application and
testsuite, but don't comment on the relation between grammar and
testsuite.

\section{Conclusion}
	\lbl{ch:conclu}

The approach presented tries to make available the linguistic knowledge
that went into the grammar for development of testsuites. Grammar
development and testsuite compilation are seen as complementary and
interacting processes, not as isolated modules. We have seen that even
large testsuites cover only a fraction of existing large-coverage grammars,
and presented evidence that there is a considerable amount of redundancy
within existing testsuites.

To empirically validate that the procedures outlined above improve grammar
and testsuite, careful grammar development is required. Based on the
information derived from parsing with instrumented grammars, the changes
and their effects need to be evaluated. In addition to this empirical work,
instrumentation can be applied to other areas in Grammar Engineering, e.g.,
to detect sources of spurious ambiguities, to select sample sentences
relying on a disjunct for documentation, or to assist in the construction
of additional test cases. Methodological work is also required for the
definition of a practical and intuitive criterion to measure limited
interaction coverage.

Each existing grammar development environment undoubtely offers at least
some basic tools for comparing the grammar's coverage with a
testsuite. Regrettably, these tools are seldomly presented publicly (which
accounts for the short list of such references). It is my belief that the
thorough discussion of such infrastructure items (tools and methods) is of
more immediate importance to the quality of the lingware than the
discussion of open linguistic problems. 


\bibliographystyle{acl}
\bibliography{ref}

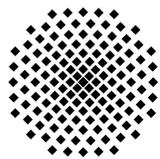
\end{document}